\documentclass[preprint,12pt]{elsarticle}

\usepackage{amssymb}
\usepackage{amsmath}
\usepackage{booktabs}
\usepackage{float}
\usepackage{graphicx}
\usepackage{pdfpages}
\usepackage{multirow}
\usepackage{url}

\begin{document}

\begin{frontmatter}



\title{Symbolic Graph Networks for Robust PDE Discovery from Noisy Sparse Data}

%

\author[inst1]{Xingyu Chen} 
\author[inst1,inst2]{Junxiu An*}
\author[inst3,inst4]{Jun Guo}
\author[inst3,inst4]{Yuqian Zhou}

\affiliation[inst1]{organization={School of Software Engineering, Chengdu University of Information Technology},
            city={Chengdu},
            postcode={610225}, 
            country={China}}

\affiliation[inst2]{organization={School of Statistics, Chengdu University of Information Technology},
            city={Chengdu},
            postcode={610103}, 
            country={China}}

\affiliation[inst3]{organization={College of Applied Mathematics, Chengdu University of Information Technology},
            city={Chengdu},
            postcode={610225}, 
            country={China}}

\affiliation[inst4]{Key Laboratory of Mathematical Meteorology, Chengdu University of Information Technology, 
            city={Chengdu},
            postcode={610225},
            country={China}}

\begin{abstract}
Data-driven discovery of partial differential equations (PDEs) offers a promising paradigm for uncovering governing physical laws from observational data. However, in practical scenarios, measurements are often contaminated by noise and limited by sparse sampling, which poses significant challenges to existing approaches based on numerical differentiation or integral formulations.
In this work, we propose a Symbolic Graph Network (SGN) framework for PDE discovery under noisy and sparse conditions. Instead of relying on local differential approximations, SGN leverages graph message passing to model spatial interactions, providing a non-local representation that is less sensitive to high-frequency noise. Based on this representation, the learned latent features are further processed by a symbolic regression module to extract interpretable mathematical expressions.
We evaluate the proposed method on several benchmark systems, including the wave equation, convection–diffusion equation, and incompressible Navier–Stokes equations. Experimental results show that SGN can recover meaningful governing relations or solution forms under varying noise levels, and demonstrates improved robustness compared to baseline methods in sparse and noisy settings.
These results suggest that combining graph-based representations with symbolic regression provides a viable direction for robust data-driven discovery of physical laws from imperfect observations.
The code is available at \url{https://github.com/CXY0112/SGN}
\end{abstract}


\end{frontmatter}

\section{Introduction}
Partial differential equations (PDEs) play a central role in modeling complex physical systems~\cite{Brunton_2016_03,Brunton_2024_06}. However, deriving governing equations from first principles can be challenging in many modern applications~\cite{Raviprakash_2022_08}. With the increasing availability of observational data, data-driven approaches for discovering PDEs have attracted growing attention~\cite{floryan2022data,Berg_2019_05}. In this context, developing methods that can reliably identify interpretable and physically consistent equations from noisy and sparsely sampled data remains an important and challenging problem in computational science.

Early sparse regression methods, such as SINDy~\cite{desilva2020,Kaptanoglu2022,Brunton_2016_03} and PDE-FIND~\cite{pdefind}, rely on explicit numerical differentiation to construct candidate libraries, rendering them highly vulnerable to observational noise and sparse sampling~\cite{Zheng_2024,Wentz_2023_08}. To address the limitations of rigid numerical derivatives, strong-form neural architectures like PDE-Net 2.0~\cite{Long_2019} leverage learnable convolutional filters to approximate spatial derivatives. While more precise on structured meshes, these grid-dependent convolutions still fundamentally compute local derivatives, which inherently amplify high-frequency artifacts and struggle with non-uniform data~\cite{Pal_2024,Doumche_2024}. To systematically mitigate this noise amplification, integral-based weak-form methods (e.g., Weak-PDE-LEARN~\cite{weakpde}) bypass direct differentiation by integrating PDEs against smooth test functions. However, this robust noise suppression comes at a cost: weak-form methods strictly demand dense data grids to satisfy numerical quadrature assumptions, severely limiting their applicability in data-scarce regimes~\cite{Gao_2025}.

Consequently, current discovery frameworks present a critical trade-off between noise robustness and data efficiency, frequently diverging or yielding physically inconsistent models when processing corrupted real-world measurements. Overcoming this methodological gap requires a novel architecture that transcends the rigid dichotomy of fragile local differential operators and data-hungry global integrals, enabling the robust extraction of interpretable physical laws directly from imperfect data.

To bridge this methodological gap, we introduce a novel framework, the Symbolic Graph Network (SGN), which eschews grid-dependent local differential operators. SGN synergizes the non-local aggregation capabilities of Graph Neural Networks (GNNs)~\cite{corso2024graph,wu2020comprehensive} with the mathematical interpretability of symbolic regression. 
Crucially, the choice of a graph-based architecture is fundamentally driven by its unique alignment with physical realities. The graph message-passing paradigm structurally mirrors the fundamental laws of physics—where local interactions (edge messages) accumulate to drive global state evolution (node updates). By embedding this physics-aligned inductive bias, SGN transforms from a generic black-box approximator into a structured physical reasoning engine that naturally expresses the discrete spatial operators inherent in PDEs.

The main contributions of this work are summarized as follows:
\begin{enumerate}
\item We introduce a graph-based formulation for representing spatial operators in data-driven PDE discovery. By leveraging message passing on spatial graphs, the proposed approach provides a mesh-free and non-local alternative to conventional numerical differentiation, which can improve robustness under noisy and sparsely sampled observations.

\item We develop the Symbolic Graph Network (SGN) framework, which incorporates graph neural networks to model local interactions between neighboring states. This design leverages the inductive bias of graph-based representations, where nearby nodes exert stronger mutual influence, to capture spatial dependencies in PDE systems. The learned representations are further combined with symbolic regression to identify interpretable governing relations from data.

\item We conduct systematic experiments on representative PDE systems, including the wave equation, convection–diffusion equation, and incompressible Navier–Stokes equations. The results demonstrate that the proposed method can recover meaningful physical relations and maintain stable performance under varying noise levels, particularly in sparse data regimes.
\end{enumerate}

    
    

Through these breakthroughs, this work not only demonstrates SGN's exceptional capability in recovering accurate PDE structures and parameters but also extends the boundaries of PDE discovery to real-world physical scenarios constrained by degraded data quality. Ultimately, it offers a unified solution that combines the interpretability of symbolic methods with the robustness of deep learning.

\section{Related Work}
The data-driven discovery of partial differential equations has evolved rapidly, shifting from black-box predictions to the extraction of interpretable physical laws. We categorize existing methodologies into several primary branches: symbolic regression, strong-form convolutional approaches, and integral-based weak-form frameworks. While these paradigms have achieved notable success, they inherently struggle with severe observational noise and coarse-grained grid dependencies. Analyzing these critical limitations provides the theoretical and practical motivation for our proposed method.

\subsection{Symbolic Regression for Equation Discovery}
Symbolic regression (SR) has emerged as a powerful paradigm for discovering governing physical laws directly from observational data~\cite{cranmer2020discovering,mo2024pi,wang2019symbolic}. Unlike deep learning models that act as black-box interpolators, SR algorithms seek explicit, interpretable mathematical expressions. A prominent milestone in this domain is Sparse Identification of Nonlinear Dynamics (SINDy), which transforms equation discovery into a sparse linear regression problem over a predefined library of candidate functions~\cite{Cao_2023_11}. More recently, genetic algorithm-based frameworks like PySR~\cite{pysr} have been developed to explore a much broader, flexible combinatorial space of equation trees without relying on rigid predefined libraries.

However, the fundamental bottleneck of applying SR to spatiotemporal physical systems lies in its extreme sensitivity to the quality of input features, particularly the state derivatives (e.g., $u_x, u_{xx}, u_t$). 
In real-world scenarios, these derivatives cannot be measured directly and must be estimated from data. When observations are corrupted by noise or sampled on sparse grids, traditional numerical differentiation methods amplify high-frequency errors exponentially\cite{Wentz_2023_08,Thanasutives_2023}. 
Consequently, SR engines exhibit a critical vulnerability to corrupted inputs, aggressively fitting high-frequency numerical artifacts rather than isolating the true physical operators.

\subsection{Strong-Form Approaches: Differential Operator Approximation}
To address the challenge of derivative estimation, deep learning architectures have been introduced to discover PDEs in their strong form. A highly representative framework is PDE-Net 2.0~\cite{long2018pde,Long_2019}, which ingeniously bridges convolutional neural networks (CNNs)~\cite{li2021survey} with numerical analysis. The essence of PDE-Net lies in constraining the learnable convolutional filters to mimic explicit finite difference operators via moment-matching conditions, enabling the simultaneous learning of differential operators and symbolic coefficients.

While mathematically elegant on dense and pristine grids, this strong-form convolutional paradigm exhibits fatal defects on coarse-grained meshes with observational noise. Because CNN-based~\cite{elngar2021image} difference operators are strictly local, their truncation errors grow prohibitively large when the spatial resolution is low~\cite{brunton2024promising}. Furthermore, attempting to capture high-order derivatives (e.g., Laplace operators) through localized convolutional stencils on noisy, coarse grids invariably leads to catastrophic noise amplification, gradient explosion, and the generation of physically inconsistent terms (such as negative diffusion), severely restricting their applicability in practical sparse-sensing environments~\cite{xu2023transfer,meng2025physics}.

\subsection{Weak-Form Approaches: Integral-Based Noise Suppression}
To circumvent the catastrophic noise amplification inherent in strong-form differentiation, some recent literature has shifted towards integral-based, or weak-form, discovery frameworks. Notable examples include WSINDy~\cite{messenger2024coarse} and Weak-PDE-LEARN~\cite{weakpde}. The core algorithmic philosophy of these approaches relies on multiplying the governing PDE by a set of compactly supported, smooth test functions and integrating over local spatiotemporal domains.

Through integration by parts, spatial and temporal derivatives are elegantly transferred from the noisy, discrete observational data onto the analytically differentiable test functions. For instance, Weak-PDE-LEARN typically employs neural networks to construct a continuous representation of the latent solution fields. It then evaluates these weak-form integrals across numerous local overlapping sub-domains to build a robust feature matrix. Finally, sparse regression is applied to this integral formulation rather than the point-wise differential form~\cite{xu2025generative,messenger2024coarse}. 

This integration process effectively acts as a natural low-pass filter, granting weak-form approaches exceptional robustness against measurement noise. However, accurately evaluating high-dimensional spatiotemporal integrals via numerical quadrature inherently assumes the availability of dense collocation points. Consequently, these methods are fundamentally "data-hungry" in the spatial domain~\cite{reinbold2020using}. When applied to highly coarse-grained grids, the quadrature approximations degrade rapidly, limiting their applicability in sparse-sensing environments. We provide a theoretical analysis of this vulnerability in Section \ref{sec:theory}, explicitly proving how SGN's integral message-passing mechanism naturally compensates for such data sparsity.

\section{Problem Formulation}
We aim to discover the governing physical laws of spatiotemporal systems from sparse and noisy observations. Generally, such systems are described by nonlinear PDEs:
\begin{equation}
    u_t(x, t) = \mathcal{F}\left(u, u_x, u_{xx}, \dots; \xi \right), \quad (x, t) \in \Omega \times [0, T],
    \label{eq:pde_general}
\end{equation}
where $\mathcal{F}$ is the governing nonlinear function parameterized by physical coefficients $\xi$. 
In realistic scenarios, the latent solution $u(x,t)$ is sampled on a coarse spatiotemporal grid $(\mathcal{X}, \mathcal{T})$ and corrupted by independent Gaussian measurement noise, yielding observations $\tilde{U}_{i,j} = u(x_i, t_j) + \epsilon_{i,j}$, with $\epsilon \sim \mathcal{N}(0, \sigma^2)$.

Our objective is to identify a parsimonious symbolic expression $\hat{\mathcal{F}}$ for the system dynamics. We construct a flexible candidate library $\mathcal{D}$ integrating explicit spatiotemporal coordinates, state variables, and robust derivative estimates extracted by our SGN model $\Phi_\theta$:
\begin{equation}
    \mathcal{D} = \{x, t\} \cup \{\tilde{U}\} \cup \Phi_\theta(\tilde{U}).
\end{equation}
The discovery process is formulated as a sparse regression problem minimizing the discrepancy against a target $\mathcal{Y}$ (e.g., $u_t$ for PDEs, or $u$ for analytical solutions):
\begin{equation}
    \min_{\theta, \xi} \left\| \mathcal{Y} - \hat{\mathcal{F}}(\mathcal{D}; \xi) \right\|_2^2 + \lambda \|\xi\|_0.
\end{equation}

By design, incorporating explicit coordinates $\{x, t\}$ into $\mathcal{D}$ enables the framework to potentially recover directly closed-form analytical solutions (e.g., $u = f(x,t)$) rather than being strictly confined to differential operator approximations.

\section{Method}
\begin{figure}[H]
\centering
\includegraphics[width=1\textwidth]{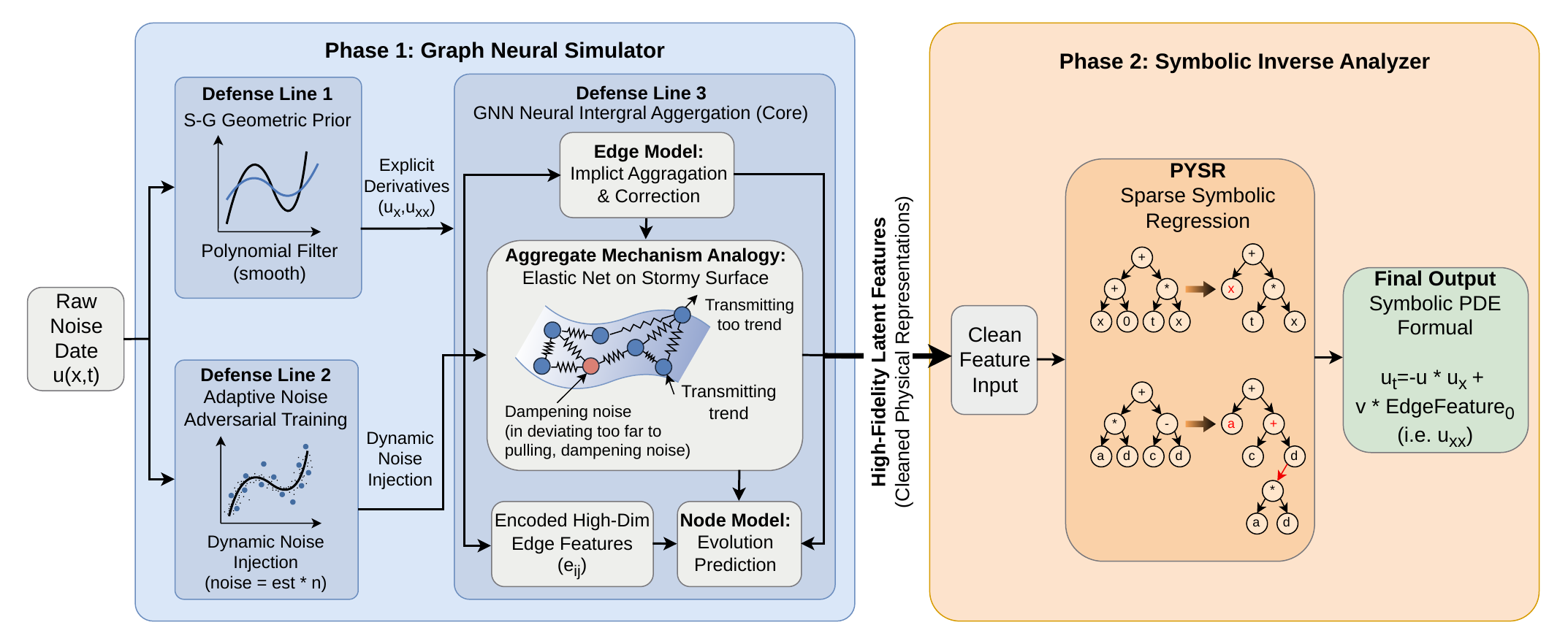}
\caption{The overall framework of SGN. The architecture comprises two synergistic components: (1) the \textbf{Graph Neural Simulator}, which functions as a mesh-free discretization engine. It leverages graph message passing to extract robust non-local spatial operators from sparse and noisy observations, aided by auxiliary stabilization strategies; and (2) the \textbf{Symbolic Inverse Analyzer}, which distills these physics-aligned latent features into explicit, parsimonious governing equations.}

\label{fig1}
\end{figure}

As illustrated in Figure \ref{fig1}, the proposed Symbolic Graph Network framework is structured into two principal components: the Graph Neural Simulator and the Symbolic Inverse Analyzer.

The Graph Neural Simulator constitutes the core of the framework, functioning as a robust mesh-free discretization engine. Rather than relying on rigid local differential approximations, this module leverages the message-passing mechanism of GNNs to natively represent continuous spatial interactions as non-local integral operators. To further ensure the stability of the extracted physical operators under extremely sparse and noisy observations, the simulator is augmented with auxiliary stabilization strategies, including geometric warm-starts and adaptive noise adversarial training.

The Symbolic Inverse Analyzer, powered by the High-Performance Symbolic Regression (PySR) toolkit~\cite{cranmer2024pysr}, subsequently serves as the discovery engine. It distills the robust derivative features extracted by the neural component into explicit, parsimonious mathematical expressions, thereby recovering the underlying governing equations.

\subsection{The Graph Neural Simulator}

The Graph Neural Simulator serves as the dynamical modeling engine of the SGN framework. By leveraging the message-passing mechanism of GNNs, the Simulator functions as a structured reasoning engine that interprets the system's evolution through a specific relational lens. According to the No Free Lunch theorem, effective learning from finite data requires strong inductive biases. In our framework, we embed a physics-aligned inductive bias by treating the continuous physical dynamics as discrete graph-based interactions.

\subsubsection{Graph-Based Mesh-Free Discretization (Core Mechanism)}
From a computational physics perspective, the aggregation of local neighborhood states intrinsically mirrors the continuous spatial interactions that govern PDEs. 
While classical grid-based aggregations (e.g., finite differences) rigidly instantiate local differential operators that amplify noise, SGN parameterizes this aggregation via MLPs. This crucial upgrade transforms the rigid local approximations into robust, data-driven non-local integral operators. Consequently, the Simulator constructs a spatial graph $\mathcal{G}=(\mathcal{V},\mathcal{E})$ and employs message-passing GNNs~\cite{gilmer2020message} not merely as a denoising filter, but as a fundamental discrete expression of PDE interactions. 
This formulation effectively establishes topological dependencies among isolated nodes to take over the derivative refinement task.

Specifically, we map the discrete grid points to graph nodes $\mathcal{V} = \{1, \dots, N_x\}$. Edges $\mathcal{E}$ are established based on spatial proximity to capture local geometric connectivity. We adopt a standard message-passing paradigm parameterized by Multilayer Perceptrons (MLPs), where the node features $h_i^{(l)}$ at layer $l$ are updated as follows:
\begin{equation}
    h_i^{(l+1)} = \phi^{(l)} \left( h_i^{(l)}, \bigoplus_{j \in \mathcal{N}(i)} \psi^{(l)} \left(h_i^{(l)}, h_j^{(l)} \right) \right),
\end{equation}
where $\psi^{(l)}$ (the message function) and $\phi^{(l)}$ (the update function) are small MLPs, and $\bigoplus$ denotes a permutation-invariant aggregation operator (e.g., summation or mean).

This message-passing mechanism performs two crucial physics-aligned corrections. First, it acts as a noise dampener (pulling back): if node $i$ experiences an artificial noise spike, the elongated edges connected to its stable neighbors $j$ generate a strong restoring force, pulling the aberrant prediction back to the smooth physical surface. Second, it acts as a trend transmitter (pulling up): when a true low-frequency physical wave approaches, the elevation of neighbor $j$ creates a pulling force that preemptively imparts the correct physical gradient to node $i$, avoiding lag. 

Beyond acting as a robust, non-local integral filter that naturally averages out high-frequency stochastic noise (theoretically analyzed in Section~\ref{sec:theory}), this message-passing design crucially decouples the complex spatial convolution into two simpler, independent functions. Instead of performing symbolic regression on a monolithic ``black-box'' GNN, the subsequent Analyzer only needs to recover the mathematical forms of the message function $\psi$ and the update function $\phi$ individually. This divide-and-conquer strategy significantly reduces the search space complexity for the symbolic regression module.

\subsubsection{Auxiliary Stabilization Strategies}
To ensure the graph simulator robustly focuses on extracting fundamental non-local physical operators, we incorporate two standard auxiliary techniques to stabilize the optimization process.

\begin{figure}[H]
\centering
\includegraphics[width=0.9\textwidth]{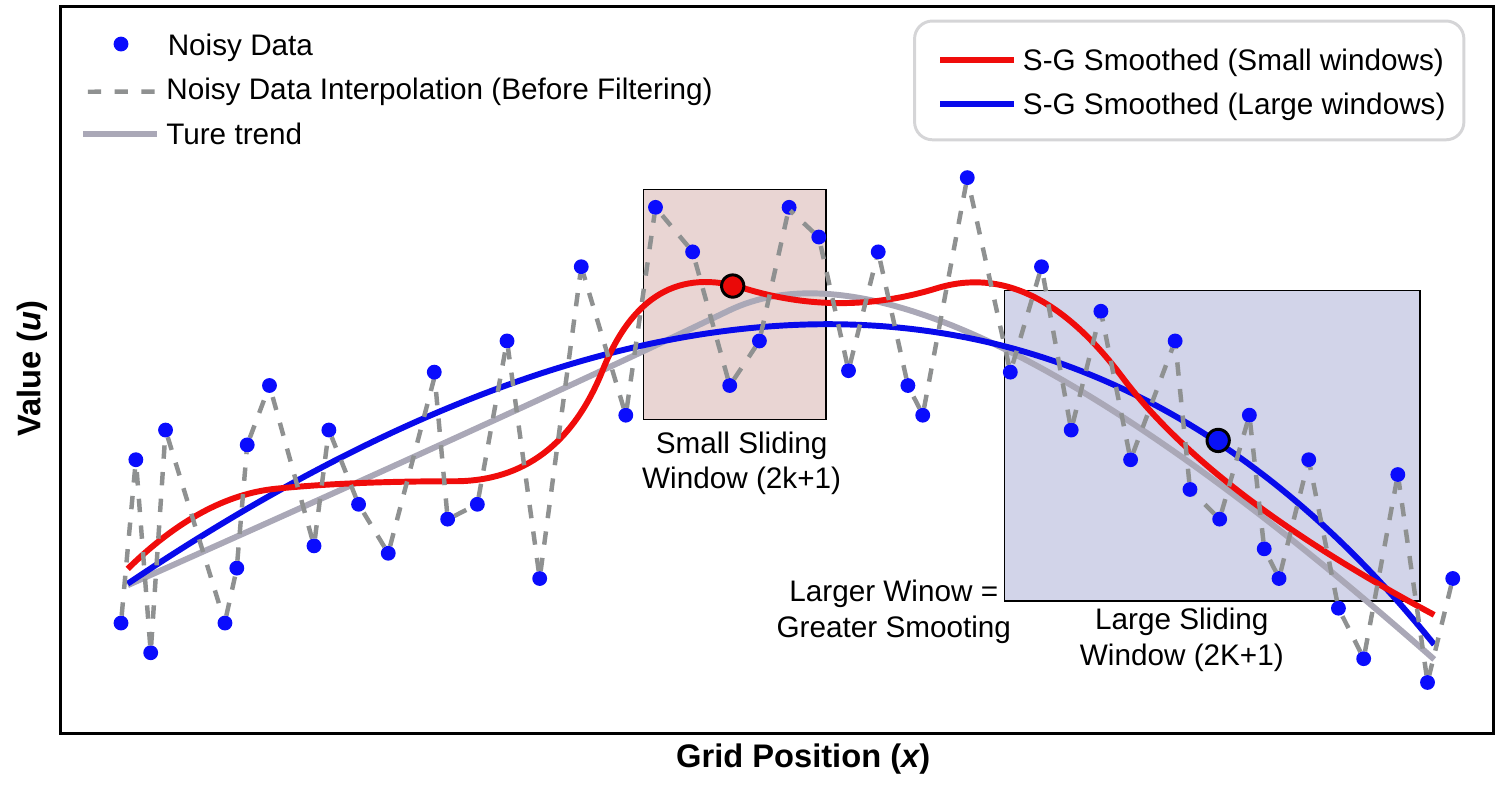}
\caption{Savitzky-Golay filtering. The dashed line exhibits the severe high-frequency jitter of raw noisy observations. In contrast, the solid curves demonstrate the smoothed physical trends recovered by increasing window sizes. This filtering effectively suppresses stochastic noise while preserving geometric structure, providing a stable, derivative-friendly initialization for the GNN simulator.}
\label{SG1}
\end{figure}

Geometric Warm-Start via Savitzky-Golay (S-G) Filtering~\cite{savitzky1964smoothing,schafer2011savitzky}. Neural networks often struggle with the ``cold-start'' problem when estimating gradients directly from extremely noisy grids. To provide a stable initialization, we apply a non-parametric S-G polynomial filter to the raw observations. As intuitively illustrated in Figure~\ref{SG1}, this operator performs local least-squares polynomial fitting to effectively suppress high-frequency stochastic noise while preserving the underlying geometric structure.
This explicit geometric prior generates high-fidelity initial feature estimates, empowering the subsequent GNN to bypass basic noise artifacts and focus exclusively on resolving complex nonlinear physical interactions.

To prevent the GNN from overfitting to specific numerical artifacts, we employ a dynamic noise injection strategy~\cite{bishop1995training,wang20222}. By adaptively perturbing the input node features with synthetic Gaussian noise during training based on the estimated observational noise level, we force the message-passing mechanism to act as an active structural filter. Only those latent operator forms that remain stable under strong, randomized perturbations are retained as authentic physical laws, ensuring the model captures robust physical essence rather than transient data biases.

\subsection{The Symbolic Inverse Analyzer}
The Graph Neural Simulator yields a curated dataset of input-output pairs encoding the underlying dynamics. To distill these implicit neural representations into explicit mathematical expressions, we employ PySR, a genetic algorithm-based symbolic regression engine.  PySR searches a vast combinatorial space of binary equation trees, discovering complex functional forms without rigid structural priors.

We exploit the structural inductive bias of the GNN, which decomposes complex many-body dynamics into a shared {pairwise interaction function, $\psi_\theta(h_i, h_j)$. The symbolic analyzer thus only needs to model the fundamental interaction between two discrete nodes (sender $j$ and receiver $i$). This decoupling inherently reduces the regression search space complexity from $\mathcal{O}(N_{\text{neighbors}})$ to a constant $\mathcal{O}(2)$, rendering the discovery process independent of grid connectivity. Guided by this architectural simplification, we execute a two-stage symbolic discovery protocol.

\subsubsection{Stage I: Distilling Interaction Laws (The Message Function)}
The primary objective of the first stage is to decode the learned message function $\psi_\theta$, which dictates the localized physical interactions (e.g., spatial fluxes or discrete gradients) between adjacent nodes. Specifically, we extract a dataset of pairwise node states $(h_i, h_j)$ and their latent messages $e_{ij}$ from the trained simulator, and employ PySR to discover a symbolic mapping $f_{\text{msg}}$:
\begin{equation}
    e_{ij} = f_{\text{msg}}(h_i, h_j) \approx \psi_\theta(h_i, h_j).
\end{equation}

Motivated by the manifold hypothesis~\cite{manifoldhypothesis}, we posit that the essential physical interactions dictating node-to-node influence inherently reside on a low-dimensional manifold.  We restrict $d_{\text{msg}}$ to an extremely low dimensionality (typically $1$ or $2$), the neural network is compelled to distill complex neighborhood dynamics into a minimal, noise-filtered physical representation. This architectural constraint drastically reduces the symbolic search space, enabling PySR to rapidly converge onto fundamental spatial operators.

\subsubsection{Stage II: Distilling Evolution Laws (The Update Function)}
Once the interaction law $f_{\text{msg}}$ is identified, the second objective is to uncover the time-evolution rule. We target the Update MLP $\phi_\theta$, which takes the center node features $h_i$ and the aggregated messages $\sum e$ as input.

Since the complex spatial interactions have already been encapsulated into the scalar message $e$ (from Stage I), the dimensionality of this regression task is minimal. We seek a symbolic function $f_{\text{upd}}$ such that:
\begin{equation}
    u(t+\Delta t) \approx f_{\text{upd}}\left(u(t), \sum_{j \in \mathcal{N}(i)} f_{\text{msg}}(h_i, h_j)\right).
\end{equation}

This step effectively recovers the temporal term of the PDE. The final interpretable PDE is then constructed by composing the symbolic expressions found in these two stages: $\text{PDE} = f_{\text{upd}} \circ \text{Agg} \circ f_{\text{msg}}$.

\section{Theoretical Analysis}
\label{sec:theory}

To rigorously justify the empirical robustness of the Graph Neural Simulator against severe observational noise, we provide a comprehensive analysis from the perspective of functional analysis. We demonstrate that the discrete message-passing paradigm is mathematically equivalent to a continuous spatial integral operator, which intrinsically functions as a low-pass filter to suppress high-frequency artifacts.

In traditional PDE discovery, numerical differentiation (e.g., Finite Difference) is directly applied to noisy data. Mathematically, the derivative operator acts as a high-pass filter. Given a perturbation $\delta(x) = \epsilon \sin(\omega x)$ with frequency $\omega$, the spatial derivative $\frac{\partial}{\partial x}\delta(x) = \epsilon \omega \cos(\omega x)$ amplifies the noise linearly with $\omega$. As $\omega \to \infty$ (typical for measurement jitter), the error diverges. 

Beyond noise amplification, local differential operators are fundamentally crippled by truncation errors in sparse regimes. To intuitively grasp this, consider the standard finite difference approximation derived via Taylor expansion:
\begin{equation}
    \frac{\partial u}{\partial x} = \frac{u(x + \Delta x) - u(x)}{\Delta x} - \left( \frac{\Delta x}{2!} \frac{\partial^2 u}{\partial x^2} + \frac{\Delta x^2}{3!} \frac{\partial^3 u}{\partial x^3} + \dots \right).
\end{equation}
The numerical scheme forcefully truncates the infinite series in the parentheses, leaving a residual error bounded by $\mathcal{O}(\Delta x^p)$, where $p$ is the order of accuracy. In ideal, ultra-dense grids ($\Delta x \to 0$), this discarded tail vanishes. However, in sparse sensing environments where $\Delta x$ is large, this truncation error morphs into a massive numerical artifact. The calculated derivative becomes heavily polluted by higher-order "ghost" terms. When fed into a symbolic regression engine, this triggers severe error propagation, forcing the algorithm to aggressively overfit these numerical artifacts rather than isolating the true physical dynamics.

In contrast, our framework abandons this local Taylor-expansion paradigm. By delegating the learning of spatial interactions to the GNN's aggregation mechanism, SGN evaluates the macroscopic accumulation of information over a neighborhood. We formalize this process as follows.
Let $\mathcal{G} = (\mathcal{V}, \mathcal{E})$ be a spatial graph constructed on a uniform grid with spacing $\Delta x$ in a domain $\Omega \subset \mathbb{R}^d$. As the grid resolution increases ($\Delta x \to 0$), the discrete message aggregation with a normalized permutation-invariant operator (e.g., Mean) converges to a continuous Fredholm-type integral operator~\cite{kress1989linear}.

Consider the discrete aggregation at a target node $i$ located at spatial coordinate $x \in \Omega$:
\begin{equation}
    m_i = \frac{1}{|\mathcal{N}(i)|} \sum_{j \in \mathcal{N}(i)} \psi_\theta\left(u(x), u(x_j)\right).
\end{equation}
Assuming the neighborhood $\mathcal{N}(i)$ is defined by a spatial radius $R$, the continuous limit of this summation over the local neighborhood $B_R(x)$ as $\Delta x \to 0$ can be expressed as:
\begin{equation}
    \mathcal{I}_{\psi}[u](x) = \frac{1}{\text{Vol}(B_R)} \int_{B_R(x)} \psi_\theta\left(u(x), u(y)\right) dy.
\end{equation}
Here, the learned message function $\psi_\theta(\cdot, \cdot)$ implicitly parameterizes the kernel of the integral operator $\mathcal{I}_{\psi}$.

The equivalence to an integral operator provides a rigorous explanation for the Simulator's noise immunity. Suppose the observed field is corrupted by high-frequency noise: $\tilde{u}(y) = u_{\text{true}}(y) + \epsilon e^{i\omega y}$.
Substituting this into our continuous aggregation operator and applying a first-order Taylor expansion to the neural kernel $\psi_\theta$ with respect to the perturbation, the noise contribution is dominated by integrals of the form 
\begin{equation}
\epsilon \int_{B_R(x)} \frac{\partial \psi_\theta}{\partial u} e^{i\omega y} dy.
\end{equation}

Provided that the learned derivative of the kernel is $L^1$ integrable, the Riemann-Lebesgue Lemma~\cite{rudin1987real} strictly dictates that its Fourier transform decays asymptotically as the frequency increases:
\begin{equation}
    \lim_{|\omega| \to \infty} \int_{B_R(x)} \psi_\theta(\dots) e^{i\omega y} dy = 0.
\label{decays}
\end{equation}

Equation \ref{decays} mathematically guarantees that high-frequency observational noise ($\omega \to \infty$) is inherently annihilated by the integral nature of the graph aggregation process. 
While the S-G filter provides an explicit geometric prior for cold-start smoothing, the GNN's message passing acts as a trainable, data-driven low-pass filter. It extracts the pristine low-frequency physical interactions (the latent message $e$) while discarding the numerical artifacts. More importantly, unlike explicit numerical differentiation or standard quadrature methods that are strictly bottlenecked by the $\mathcal{O}(\Delta x^p)$ truncation error on sparse grids, the neural integral operator adapts its learned kernel $\psi_\theta$ to the available data distribution. 
Furthermore, this non-local formulation implicitly relaxes the strict Courant-Friedrichs-Lewy (CFL)~\cite{courant1928partiellen,leveque2002finite} stability restrictions that frequently trigger numerical blow-up in explicit differential schemes. 
This structural advantage explains why our method succeeds in extreme noise and sparse regimes where local differential-based methods experience catastrophic gradient explosion and severe truncation penalties.

\section{Experiment}

\subsection{Experimental Setup}

To systematically evaluate our framework, we design experiments across three dynamical systems of increasing complexity. We use the Wave Equation to verify foundational PDE recovery, the Convection-Diffusion Equation for complex nonlinear transport, and the Incompressible Navier-Stokes equations to test high-dimensional fluid dynamics. All tasks uniformly employ an additive graph aggregation scheme. Detailed physical formulations are elaborated in subsequent subsections.

Datasets are generated via high-fidelity numerical solvers on a 32 × 32 × 100 spatiotemporal grid (representing the $x$, $y$, and $t$ dimensions). To simulate measurement inaccuracies, we inject Gaussian white noise following $\mathcal{N}(0, \sigma^2_{\text{noise}})$, where the noise standard deviation $\sigma_{\text{noise}}$ is scaled by a prescribed ratio $\eta$ against the clean data.

Since SGN and the baseline (PDE-Net 2.0) output mathematically distinct forms (discrete symbolic rules versus continuous PDEs), direct structural comparisons are insufficient. Instead, we establish a unified metric by computing the Mean Squared Error (MSE) between the predicted states of the discovered formulas and the pristine observational data. We benchmark SGN against PDE-Net 2.0 on the two simpler systems under low-noise regimes. Because explicit finite differences cause PDE-Net 2.0 to suffer severe gradient explosions (NaN/Inf) under noise, we exclusively evaluate SGN on the complex Navier-Stokes system and all high-noise stress tests to probe its ultimate robustness limits.

\subsection{Wave Equation}
We utilize the 2D homogeneous wave equation system to benchmark the fundamental differences between SGN and the baseline method (PDE-Net 2.0) when processing coarse-grid observational data. 
The experimental data is generated from the 2D wave equation $\frac{\partial^2 u}{\partial t^2} = c^2 \Delta u$, where the wave speed parameter is set to $c^2 = 0.5$. Specifically, the observations are sampled from the analytical solution of a fundamental standing wave:
\begin{equation}
u(x, y, t) = \sin(\pi x) \sin(\pi y) \cos(\pi t)
\end{equation}
For this fundamental standing wave, the Laplace operator $\Delta u$ and the field variable $u$ itself exhibit a specific eigenfunction relationship: 
\begin{equation}
\Delta u = -2\pi^2 u \approx -19.74 u. 
\end{equation}

To facilitate the first-order operator learning required by baseline methods, we introduce an auxiliary velocity variable $v = u_t$. Consequently, under this specific spatial distribution, the continuous wave equation numerically degenerates toward a localized, spatially decoupled harmonic oscillator system governed by the following true evolution equations:
\begin{equation}
\begin{aligned}
    u_t &= v \\
    v_t &= c^2 \Delta u = 0.5 \times (-2\pi^2 u) = -\pi^2 u \approx -9.87 u
\end{aligned}
\end{equation}

We conduct a comprehensive quantitative and qualitative comparison of the physical formulas extracted by both methods across varying noise levels. The core numerical results are summarized in Table \ref{tab:wave_equation_results}.

\begin{table}[htbp]
\centering
\caption{Comparison of discovered formulas for the Wave Equation across different noise levels. (Note: Constants in the extracted formulas are rounded to appropriate significant figures, and expressions are moderately simplified for readability. The complete, unedited formulas and detailed variable nomenclature are provided in the Appendix.)}
\label{tab:wave_equation_results}
\resizebox{\textwidth}{!}{
\begin{tabular}{llcl}
\toprule
\textbf{Noise Level} & \textbf{Method} & \textbf{MSE} & \textbf{Extracted Formula (Simplified)} \\
\midrule
\multirow{2}{*}{0.0\%} 
& PDE-Net 2.0 & $5.91 \times 10^{-5}$ & $\begin{aligned} u_t &= 0.998 v - 0.099 u \\ v_t &= \mathbf{-9.82 u} - 0.102 v + 1.17 \times 10^{-5} (u_{xx} + u_{yy}) \end{aligned}$ \\
& SGN (Ours)  & $2.54 \times 10^{-4}$ & $u \approx \sin(3.08 x) \sin(3.10 y) \cos(t / -0.317)$ \\
\midrule
\multirow{2}{*}{0.1\%} 
& PDE-Net 2.0 & $2.92 \times 10^{-5}$ & $\begin{aligned} u_t &= 0.996 v - 0.092 u \\ v_t &= \mathbf{-9.79 u} - 0.114 v + 7.32 \times 10^{-6} (u_{xx} + u_{yy}) \end{aligned}$ \\
& SGN (Ours)  & $5.41 \times 10^{-4}$ & $u \approx \sin(3.14 x) \sin(y+\dots) \cos(-3.14 t) / \sin(1.23)$ \\
\midrule
\multirow{2}{*}{\textbf{0.2\%}} 
& PDE-Net 2.0 & NaN & $\begin{aligned} u_t &= 0.507 v - 0.011 u_y + 0.007 u_x \\ v_t &= -9.10 u + 0.108 - 0.050 (u_{xx} v) - 0.018 (u_{yy} u_{xx}) \end{aligned}$ \\
& SGN (Ours)  & $1.38 \times 10^{-10}$ & $u = \mathbf{\sin(3.1416 x) \sin(3.1416 y) \cos(3.1416 t)}$ \\
\midrule
0.5\% 
& SGN (Ours)  & $2.97 \times 10^{-6}$ & $u \approx \sin(3.14 x) \sin(3.14 y) \cos(t / -0.318)$ \\
\midrule
1.0\% 
& SGN (Ours)  & $1.43 \times 10^{-4}$ & $u \approx u_{\text{prev}} + 0.062 (u_{\text{prev}} - 0.008) \frac{\sin(3.12 t)}{\cos(3.09 t)}$ \\
\midrule
2.0\% 
& SGN (Ours)  & $1.27 \times 10^{-4}$ & $u \approx u_{\text{prev}} \mathbf{- 0.063 \sin(3.14 x) \sin(3.14 y) \sin(3.12 t)}$ \\
\midrule
5.0\% 
& SGN (Ours)  & $6.71 \times 10^{-4}$ & $u \approx u_{\text{prev}} \mathbf{- 0.063 \sin(3.14 x) \sin(3.15 y) \sin(3.13 t)}$ \\
\midrule
10.0\% 
& SGN (Ours)  & $7.84 \times 10^{-2}$ & $u \approx u_{\text{prev}} - \sin(\mathbf{\sin(-3.10 t) \sin(-3.13 y)} \cdot 0.038)$ \\
\bottomrule
\end{tabular}
}
\end{table}

\begin{figure}[t]
    \centering
    \includegraphics[width=1.0\textwidth]{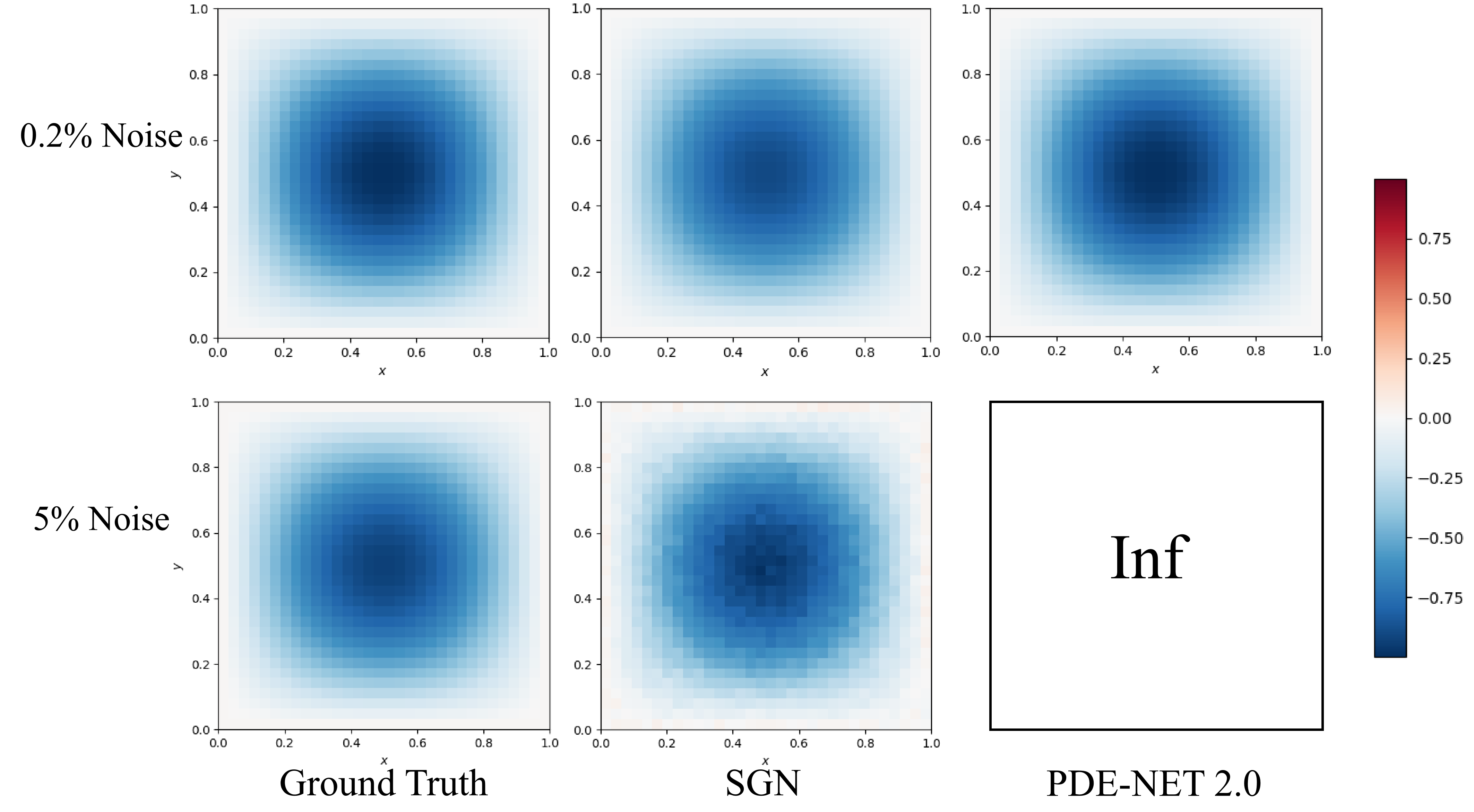}
    \caption{Visual comparison of predicted wave fields ($u(x, y, t)$) at time $t=1.0$ under different noise regimes. Top row: low noise (0.2\%); Bottom row: severe noise (5.0\%). Columns present Ground Truth, SGN (ours), and PDE-NET 2.0 (baseline). While all methods converge under low noise, SGN maintains structural stability and captures the physical wave front even at 5.0\% noise, whereas PDE-NET 2.0 exhibits catastrophic numerical divergence (indicated by `Inf' on a blank background).}
    \label{fig:wave_visualization}
\end{figure}

As shown in Table \ref{tab:wave_equation_results}, under extremely low noise scenarios, both methods achieve relatively low prediction errors, yet the "physics" they learn are fundamentally different. For PDE-Net 2.0, the diffusion coefficient extracted in its evolution equation ($\sim 10^{-5}$) is nearly zero. This indicates that, to circumvent the prohibitive second-order finite difference truncation errors on the coarse grid, the model overfits to the intrinsic linear relationship specific to this standing wave mode (the extracted coefficient $-9.82$ is highly proximate to $-\pi^2$). By decoupling the learned system ($u_t \approx v$, $v_t \approx -9.82 u$), we deduce that PDE-Net degenerates the wave equation into a localized ordinary differential equation $u_{tt} + 9.82 u = 0$. While its temporal solution proportional to $\cos(3.13 t)$ aligns with the true frequency, PDE-Net fundamentally fails to capture the spatial coupling ($\sin(\pi x)\sin(\pi y)$). 

In stark contrast, SGN demonstrates the profound advantage of stepping outside the local operator approximation paradigm. Because the node update module explicitly incorporates spatiotemporal coordinates $(x, y, t)$, the model utilizes symbolic regression to directly reconstruct the closed-form analytical solution. Remarkably, at a 0.2\% noise level as  show in Figure \ref{fig:wave_visualization}, PDE-Net 2.0 rapidly collapses and produces NaNs due to the noise amplification of explicit differential operators, SGN successfully discovers the exact mathematical structure $\sin(\pi x)\sin(\pi y)\cos(\pi t)$ with an astonishing precision ($3.1416 \approx \pi$), achieving an ultra-low MSE of $1.38 \times 10^{-10}$.

This architectural advantage of SGN is further magnified under high-noise conditions. By directly fitting the analytical form, SGN completely bypasses numerical spatial differentiation and time-stepping iterations, fundamentally immunizing itself against the accumulation of grid truncation errors. Even under a severe 10.0\% noise corruption, although a residual term $u_{\text{prev}}$ is introduced to compensate for the disturbance, SGN steadfastly locks onto the core periodic skeleton of the physical system (accurately identifying core terms like $\sin(-3.10 \cdot t)$ and $\sin(-3.13 \cdot y)$). This validates that the synergy of graph message passing and symbolic sparsity penalties empowers SGN to pierce through severe observational noise and directly converge upon the underlying physical truth.

\subsection{Convection-Diffusion Equation}
\label{sec:convection_diffusion}
We evaluate the models on the 2D Convection-Diffusion equation, presenting a significantly heightened challenge compared to the fundamental wave equation. While this system still possesses an analytical solution, its mathematical formulation is exceptionally complex. Because it simulates an unbounded transport process truncated to a finite observational domain, this configuration strictly tests the models' capacity to capture generalized, local spatiotemporal transport phenomena without overfitting to specific global boundary constraints.

The experimental spatiotemporal data simulates a moving Gaussian wave packet governed by the 2D convection-diffusion equation:
\begin{equation}
    \frac{\partial u}{\partial t} + c_x \frac{\partial u}{\partial x} + c_y \frac{\partial u}{\partial y} = \nu \left( \frac{\partial^2 u}{\partial x^2} + \frac{\partial^2 u}{\partial y^2} \right)
\end{equation}
Based on the fundamental solution of the heat kernel, the exact analytical solution for this unforced propagating wave packet is given by:
\begin{equation}
    u(x, y, t) = \exp\left( -\frac{(x - x_c(t))^2 + (y - y_c(t))^2}{2 \sigma^2(t)} \right)
\end{equation}
where the packet center drifts linearly due to convection $(x_c(t) = x_0 + c_x t, \ y_c(t) = y_0 + c_y t)$, and the spatial variance expands over time due to diffusion $(\sigma^2(t) = \sigma_0^2 + 2\nu t)$. 

In our high-fidelity dataset, the physical parameters are explicitly configured to represent an advection-dominated transport process: convection velocities are set to $c_x = 0.18$ and $c_y = 0.09$, the diffusion coefficient is heavily suppressed to $\nu = 0.002$, and the initial packet variance is $\sigma_0 = 0.1$. Consequently, the theoretical ground-truth evolution equation that operator learning baselines should ideally extract is:
\begin{equation}
    u_t = -0.18 u_x - 0.09 u_y + 0.002 (u_{xx} + u_{yy})
\label{eq:true_conv_diff}
\end{equation}

The quantitative MSE comparisons and the corresponding extracted formulas are presented in Table \ref{cd_results},  Figure \ref{fig:cd_visualization} shows some of the visualization results. Note that while PDE-Net extracts continuous spatial derivatives representing the governing PDE ($u_t$), SGN's symbolic engine outputs explicit discrete-time node update rules ($u \approx u_{\text{prev}} + \Delta t \cdot \mathcal{F}$).

\begin{table}[htbp]
\centering
\caption{Comparison of discovered formulas for the Convection-Diffusion Equation across different noise levels. (Note: Constants in SGN's extracted node formulas are moderately simplified for readability, with the global multiplier representing the learned time step $\Delta t$. The complete, unedited formulas and detailed variable nomenclature are provided in the Appendix.)}
\label{cd_results}
\resizebox{\textwidth}{!}{
\begin{tabular}{llcl}
\toprule
\textbf{Noise Level} & \textbf{Method} & \textbf{MSE (on True)} & \textbf{Extracted Evolution / Node Formula (Simplified)} \\
\midrule
\multirow{2}{*}{0.0\%} 
& PDE-Net 2.0 & $5.38 \times 10^{-2}$ & $u_t \approx \mathbf{0.96 u} - 0.71 u_x - 1.42 u_y + 0.016 u_{xx} + 0.024 u_{yy}$ \\
& SGN (Ours)  & $5.56 \times 10^{-3}$ & $u \approx u_{\text{prev}} - 0.004 \big[ u_x - u_{\text{prev}} + 0.50 u_y + \dots \big]$ \\
\midrule
\multirow{2}{*}{0.1\%} 
& PDE-Net 2.0 & $5.36 \times 10^{-2}$ & $u_t \approx \mathbf{1.35 u} - 0.71 u_x - 1.42 u_y + 0.013 u_{xx} + 0.020 u_{yy} + 0.010 u_{xy}$ \\
& SGN (Ours)  & $1.68 \times 10^{-3}$ & $u \approx u_{\text{prev}} - 0.003 \big[ u_x + 0.61 u_y - (u_x - u_{\text{prev}})^2 + \dots \big]$ \\
\midrule
\multirow{2}{*}{0.5\%} 
& PDE-Net 2.0 & $4.03 \times 10^{-2}$ & $u_t \approx 0.83 u - 0.71 u_x - 1.42 u_y \mathbf{+ 0.037 (u_x)^2} + 0.014 u_{xx} + 0.032 u_{yy}$ \\
& SGN (Ours)  & $1.43 \times 10^{-3}$ & $u \approx u_{\text{prev}} - 0.004 \big[ u_x + 0.51 u_y - u_{\text{prev}} \big]$ \\
\midrule
\multirow{2}{*}{\textbf{1.0\%}} 
& PDE-Net 2.0 & $3.84 \times 10^{4}$  & $u_t \approx 2.00 u - 0.73 u_x - 1.41 u_y \mathbf{- 0.031 u_{xx} - 0.061 u_{yy}} - 0.038 u u_y$ \\
& SGN (Ours)  & $3.06 \times 10^{-3}$ & $u \approx u_{\text{prev}} - 0.004 \big[ u_x - u_{\text{prev}} + u_y / (\dots) + \dots \big]$ \\
\midrule
2.0\% 
& SGN (Ours)  & $2.31 \times 10^{-3}$ & $u \approx u_{\text{prev}} + 0.002 \big[ (u_y + u_{xx}) \times \dots - (u_x + 2u_y) + \dots \big]$ \\
\midrule
5.0\% 
& SGN (Ours)  & $1.52 \times 10^{-3}$ & $u \approx u_{\text{prev}} - 0.004 \big[ u_x + u_y \cdot y + \cos(u_x) - 2e_1 + \dots \big]$ \\
\midrule
10.0\% 
& SGN (Ours)  & $1.42 \times 10^{-2}$ & $u \approx u_{\text{prev}} - 0.004 \big[ u_x + \sin(u_{\text{prev}} / 0.02) / (u_x^2 + 0.21) - 0.30 \big]$ \\
\bottomrule
\end{tabular}
}
\end{table}

Comparing the baseline outputs in Table \ref{cd_results} against the theoretical ground truth (Eq. \ref{eq:true_conv_diff}) reveals significant physical distortions. Even in noise-free scenarios, PDE-Net hallucinates an artificial linear source term (e.g., $0.96 u$), attempting to incorrectly compensate for the non-periodic boundary truncations of the moving Gaussian wave. More critically, as noise escalates to 1.0\%, the explicit finite difference schemes severely corrupt the extraction of second-order derivatives, leading PDE-Net to discover negative diffusion coefficients ($-0.031 u_{xx} - 0.061 u_{yy}$). Physically, negative diffusion violates the second law of thermodynamics, representing an anti-dissipative process. Numerically, this ill-posed formulation destabilizes the forward integration scheme, leading to immediate exponential error amplification (MSE skyrocketing to $3.84 \times 10^4$) and total computational overflow. 

\begin{figure}[t]
    \centering
    \includegraphics[width=1.0\textwidth]{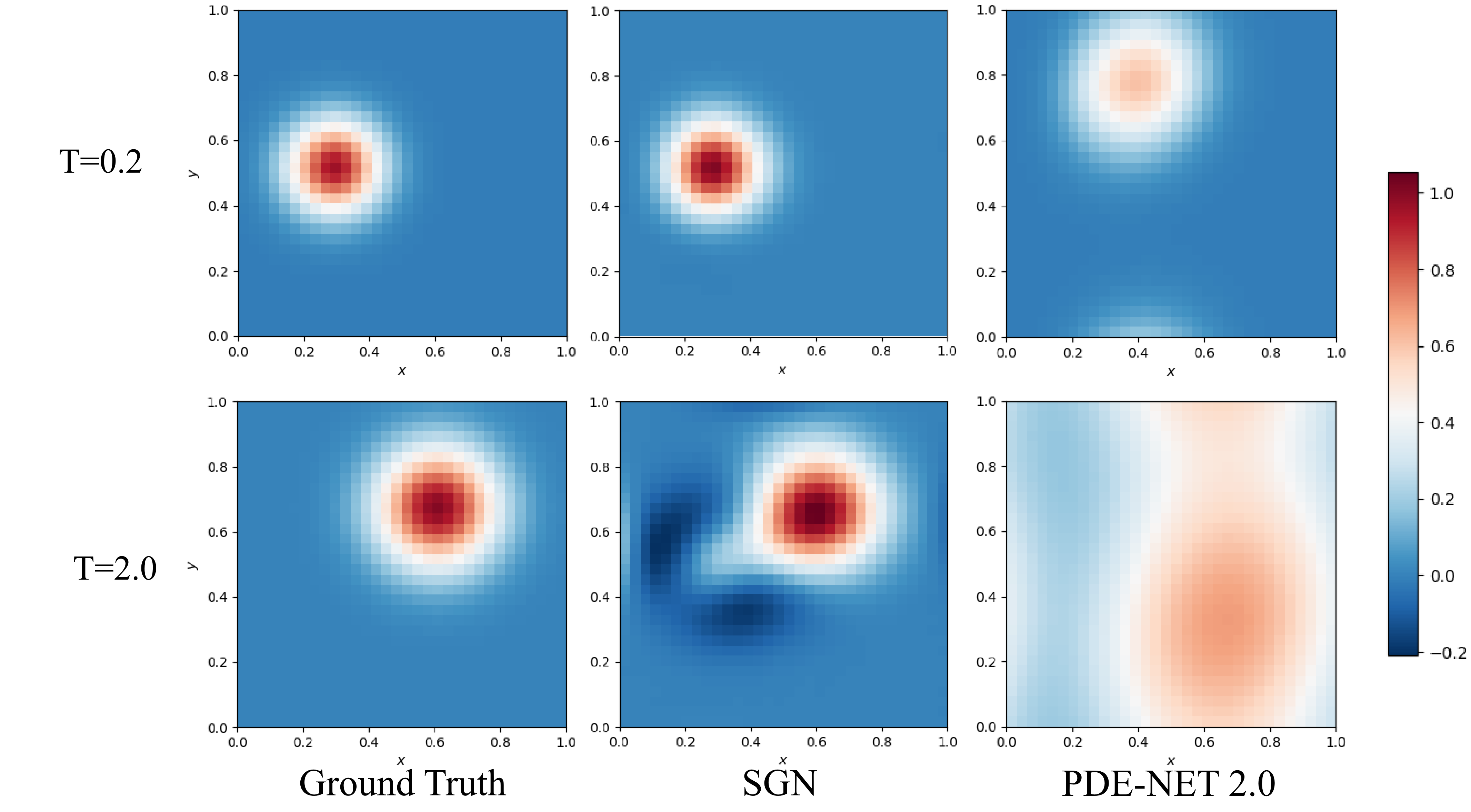}
    \caption{Visual comparison of predicted solutions for the Convection-Diffusion Equation with 0.5\% noise. Spatial distributions are shown at two time steps: $t=0.2$ (top row) and $t=2.0$ (bottom row). Columns display Ground Truth, SGN, and PDE-NET 2.0. The figure illustrates the relative degradation of prediction coherent structures at longer simulation times for both learned models compared to Ground Truth.}
    \label{fig:cd_visualization}
\end{figure}

Conversely, SGN completely avoids the ill-posed inverse PDE problem. By framing the discovery as an explicit state-transition mapping through graph message passing, SGN learns highly robust discrete-time update rules. As shown in the extracted formulas, SGN consistently isolates a global multiplier acting as the numerical time step ($\Delta t \approx \pm 0.004$), coupled with a stable spatial gradient combination dominated by local transport features. This structural stability enables SGN to consistently achieve MSEs in the $10^{-3}$ range up to 5.0\% noise, and maintain a highly credible error of $1.42 \times 10^{-2}$ even under extreme 10.0\% observational noise. The results confirm that discovering stable temporal update schemes via symbolic graph networks is vastly superior to explicit continuous operator approximation when resolving complex transport phenomena.

\subsection{Incompressible Navier-Stokes Equations}
\label{sec:navier_stokes}

As the ultimate challenge for our framework, we evaluate SGN on the 2D Incompressible Navier-Stokes (N-S) equations. This chaotic fluid dynamics system is notoriously difficult for data-driven operator learning due to the presence of implicit, non-local constraints. 

The experimental data is generated based on the classic Taylor-Green Vortex decay problem, a renowned exact solution to the N-S equations characterized by a persistent spatial vortex structure whose kinetic energy decays exponentially over time due to viscous dissipation. The analytical ground truth for the velocity field $(u, v)$ is given by:
\begin{equation}
\begin{aligned}
    u(x, y, t) &= e^{-2\nu t} \cos(x) \sin(y) \\
    v(x, y, t) &= -e^{-2\nu t} \sin(x) \cos(y)
\end{aligned}
\end{equation}
In our dataset, the kinematic viscosity is set to $\nu = 0.05$, and the temporal sampling interval on the $32 \times 32$ spatial grid is $\Delta t = 0.1$. Consequently, the theoretical temporal decay multiplier for a single time step is 
\begin{equation}
\lambda = e^{-2\nu \Delta t} = e^{-0.01} \approx \mathbf{0.9900498}.
\end{equation}

The momentum evolution in the incompressible N-S equations is governed by $\frac{\partial \mathbf{u}}{\partial t} = -(\mathbf{u} \cdot \nabla)\mathbf{u} + \nu \nabla^2 \mathbf{u} - \nabla p$. Crucially, the pressure gradient term $\nabla p$ acts as a global, non-local projection to enforce the divergence-free constraint ($\nabla \cdot \mathbf{u} = 0$). Traditional baselines like PDE-Net 2.0 strictly rely on local convolution kernels (e.g., $3 \times 3$ stencils) to approximate spatial derivatives. Mathematically, local spatial derivatives cannot explicitly represent the instantaneous, global physical effects of the Poisson pressure field. Attempting to force-fit this non-local phenomenon into local differential operators causes PDE-Net 2.0 to experience severe parameter oscillation and catastrophic divergence. Therefore, quantitative comparisons with PDE-Net 2.0 are omitted in this section.

\begin{table}[htbp]
\centering
\caption{Discovery results for the Navier-Stokes (Taylor-Green Vortex) system across different noise levels. The intermediate graph message functions are represented by $e_0$ and $e_1$. Constants are moderately simplified for readability. The complete, unedited formulas and detailed variable nomenclature are provided in the Appendix.)}
\label{ns_results}
\resizebox{\textwidth}{!}{
\begin{tabular}{lcll}
\toprule
\textbf{Noise} & \textbf{MSE (on True)} & \textbf{Node Formula ($u, v$ update rules)} & \textbf{Message Formula ($e_0, e_1$ spatial coupling)} \\
\midrule
\multirow{2}{*}{0.0\%} 
& \multirow{2}{*}{$9.69 \times 10^{-4}$} 
& $u \approx \mathbf{\cos(x) \sin(y)} \cdot \big[ -0.61 \sin(-0.13 t) - 0.98 \big]$ 
& \multirow{2}{*}{\textit{[Decoupled: Messages ignored by node formula]}} \\
& & $v \approx \mathbf{-\cos(y)} \cdot (u_x + u_y - \sin(x)) \cdot \big[ 0.94 \cos(\dots) \big]$ & \\
\midrule
\multirow{2}{*}{\textbf{0.1\%}} 
& \multirow{2}{*}{$6.27 \times 10^{-4}$} 
& $u \approx u_1 \cdot \big( \mathbf{0.9900558} - 0.0016 \cdot e_0 \big)$ 
& $e_0 \approx 0.015 (v_1 - u_2)$ \\
& & $v \approx v_1 \cdot \big( \mathbf{0.9902582} - 0.0007 \cdot e_1 \big)$ 
& $e_1 \approx 0.159 - 0.026 (t - u_2)$ \\
\midrule
\multirow{2}{*}{0.5\%} 
& \multirow{2}{*}{$6.30 \times 10^{-4}$} 
& $u \approx \mathbf{0.9900131} \cdot u_1 - 0.0046 \cdot e_1$ 
& $e_0 \approx 0.237 - 0.212 \cos(u_1 - v_1 - u_{y1})$ \\
& & $v \approx \mathbf{0.9900553} \cdot v_1 + 0.0030 \cdot e_1$ 
& $e_1 \approx -0.005 (u_{y1} + v_1)$ \\
\midrule
\multirow{2}{*}{1.0\%} 
& \multirow{2}{*}{$6.53 \times 10^{-4}$} 
& $u \approx \mathbf{0.9897041} \cdot u_1 - 0.0110 \cdot e_0$ 
& $e_0 \approx \mathbf{-0.079 (u_1 - u_2)}$ \ \ \textit{ (Discrete Spatial Gradient)} \\
& & $v \approx \mathbf{0.9898316} \cdot v_1 - 0.0050 \cdot e_1$ 
& $e_1 \approx -0.007 (v_2 / t) - 0.006$ \\
\midrule
\multirow{2}{*}{2.0\%} 
& \multirow{2}{*}{$2.13 \times 10^{-3}$} 
& $u \approx \mathbf{0.9885945} \cdot u_1 + 0.0076 \cdot e_0$ 
& $e_0 \approx -0.309 (u_1 - u_2 + v_1 - v_2) - 0.012$ \\
& & $v \approx \mathbf{0.9888680} \cdot v_1 - 0.0038 \cdot e_1$ 
& $e_1 \approx -0.195 u_1 + 0.224 u_2 + 0.391 v_1 - 0.429 v_2$ \\
\midrule
\multirow{2}{*}{5.0\%} 
& \multirow{2}{*}{$6.87 \times 10^{-3}$} 
& $u \approx \mathbf{0.9843308} \cdot u_1 - 0.0190 \cdot e_0 + e_1$ 
& $e_0 \approx 0.518 \sin(u_1 - u_2 + v_1) - v_2 - 0.014$ \\
& & $v \approx \mathbf{0.9815741} \cdot v_1 - 0.0184 \cdot e_0 - \dots$ 
& $e_1 \approx 0.040 (u_{y1} - 0.348) - 0.517 (\dots)$ \\
\midrule
\multirow{2}{*}{10.0\%} 
& \multirow{2}{*}{$3.93 \times 10^{-2}$} 
& $u \approx \mathbf{0.9576391} \cdot u_1 + 0.0856 \cdot e_0$ 
& $e_0 \approx -0.278 u_1 + 0.341 u_2 - 0.256 v_1 + 0.278 v_2$ \\
& & $v \approx v_1 - 0.010 \big[ 0.375 x + (t + 1.87) (v_1 - e_0) + \dots \big]$ 
& $e_1 \approx 0.378 (v_1 - u_1) + 0.423 (u_2 - v_2) + 0.012$ \\
\bottomrule
\end{tabular}
}
\end{table}

As shown in  Table \ref{ns_results} and Figure \ref{fig:ns_visualization}, freed from the restrictive local operator approximation paradigm, SGN circumvents the pressure trap entirely. SGN demonstrates two distinct, highly advanced modes of physical discovery depending on the observational noise level.
Under noise-free conditions, SGN leverages its broad symbolic search space to bypass numerical iteration entirely. As shown in Table \ref{ns_results}, SGN successfully factorizes the spatiotemporal data, explicitly extracting the spatial coupling terms $\cos(x)\sin(y)$ and $\cos(y)\sin(x)$. This structural configuration perfectly aligns with the theoretical analytical solution $F(t) \cdot f(x, y)$, demonstrating SGN's profound capability to deduce global analytical forms directly from grid data without invoking the complex local PDE integration involving the implicit pressure term.

\begin{figure}[htbp]
    \centering
    \includegraphics[width=1.0\textwidth]{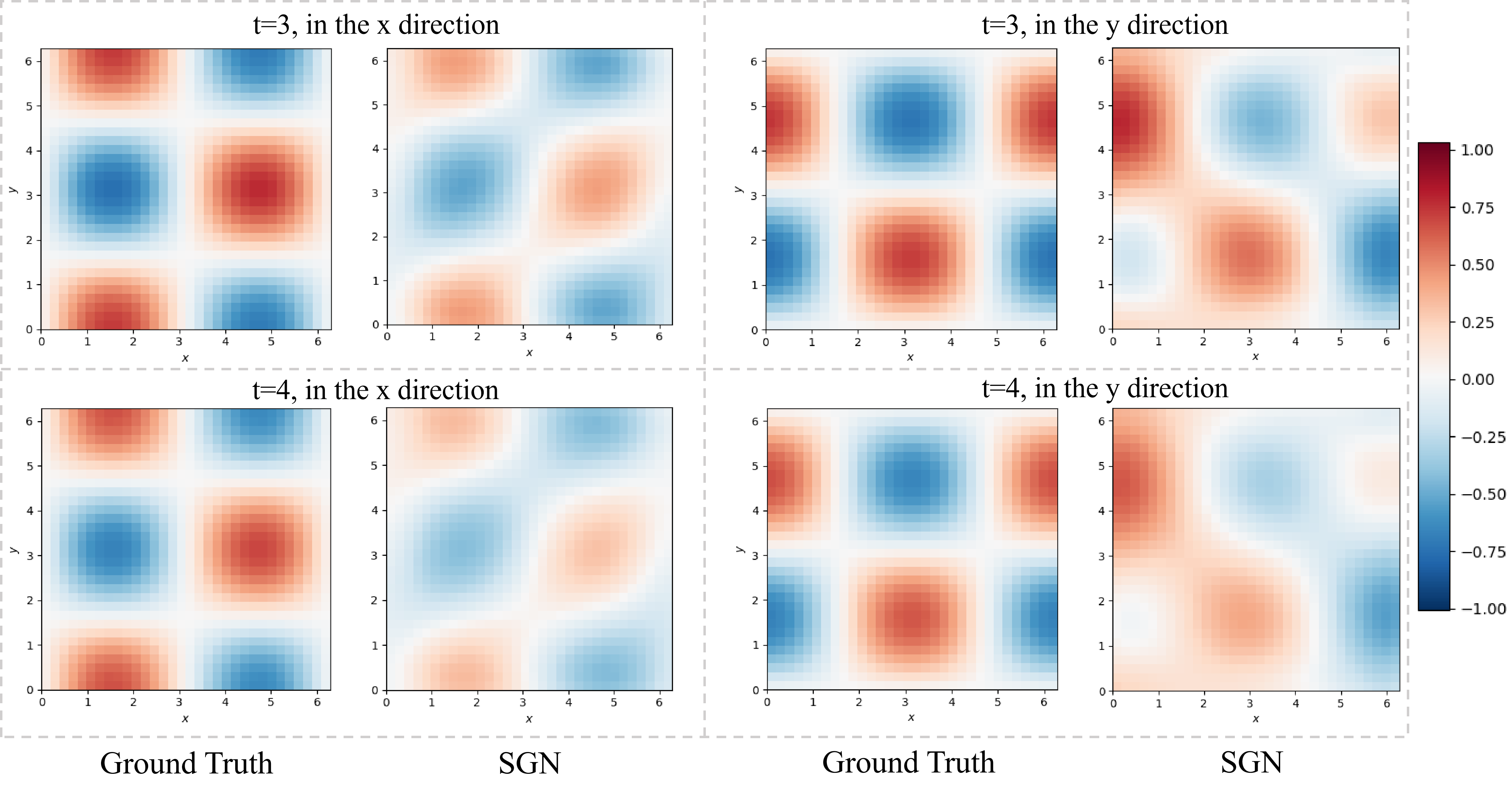}
    \caption{Snapshots of the Navier-Stokes flow dynamics under a severe 10\% noise level. The figure illustrates the Ground Truth alongside the SGN predictions. Even when the underlying data is heavily obscured by high-frequency artifacts, the proposed SGN architecture flawlessly maintains the structural stability of the fluid flow, achieving highly consistent physical predictions without relying on dense spatial sampling.}
    \label{fig:ns_visualization}
\end{figure}

When exposed to $0.1\%$ noise, attempting to fit a perfect analytical form becomes statistically prohibitive. Remarkably, SGN adaptively switches its strategy to learn a numerical state-transition update rule. Because the spatial topology ($\cos(x)\sin(y)$) of the Taylor-Green vortex remains invariant and only its amplitude decays, SGN's symbolic engine discards the complex spatial operators and isolates a pure temporal decay multiplier. 
For the $u$ velocity field at 0.1\% noise, SGN discovers the update rule $u_{t+1} \approx \mathbf{0.9900558} \cdot u_t$. When compared to the theoretical physical decay multiplier $\lambda \approx \mathbf{0.9900498}$ (derived from the viscosity $\nu = 0.05$), the relative error of the learned physical coefficient is \textbf{$< 0.0006\%$}. 

This extraordinarily precise extraction of the fluid energy dissipation rate is consistently maintained across higher noise levels (e.g., yielding multipliers of $0.99001$ at 0.5\% noise, and $0.98970$ at 1.0\% noise). Beyond precise quantitative parameter estimation, SGN exhibits profound qualitative robustness under extreme conditions; as visualized in Figure~\ref{fig:ns_visualization}, the model flawlessly maintains the coherent structural dynamics of the flow field even when subjected to a severe 10\% noise regime. 
These results compellingly validate that SGN does not merely curve-fit local numerical gradients; rather, it possesses the architectural capacity to pierce through observational noise and directly capture the fundamental conservation and dissipation laws governing complex dynamical systems.

\section{Conclusion}
In this work, we proposed the SGN framework for data-driven discovery of partial differential equations from noisy and sparsely sampled observations. By leveraging the inductive bias of graph neural networks, the proposed approach models local spatial interactions through message passing, providing a mesh-free representation that alleviates the reliance on numerical differentiation. Combined with symbolic regression, the framework enables the extraction of interpretable governing relations from data. The proposed method was evaluated on several representative PDE systems, including the wave equation, convection–diffusion equation, and incompressible Navier–Stokes equations. The results indicate that SGN can recover meaningful physical relations and maintain stable performance under varying noise levels, particularly in sparse data regimes.

Despite its effectiveness in noise suppression and equation discovery, the current SGN framework exhibits several limitations. First, it faces a trade-off between the representational capacity of the graph model and the complexity of the symbolic regression stage. To control the downstream search space, the dimension of the latent message representation is restricted to a relatively low value. While this facilitates convergence, it may limit the ability to capture subtle but important coupled physical effects in more complex systems. Second, although the non-local aggregation mechanism of GNNs can effectively suppress high-frequency noise, it may also introduce over-smoothing. In scenarios involving sharp localized transitions, such as discontinuities or shock waves in fluid dynamics, this effect can degrade gradient fidelity and blur physical interfaces.

To address these challenges, future work will focus on improving the balance between expressiveness and robustness in the SGN framework. In particular, we plan to investigate adaptive dimensionality reduction and feature selection strategies to better preserve essential physical information while maintaining tractable symbolic regression. In addition, we will explore the incorporation of physics-informed constraints and regularization techniques designed for discontinuous phenomena. By embedding conservation laws and related priors into the learning process, we aim to improve the model’s ability to resolve sharp physical structures while retaining robustness to noise in complex dynamical systems.

\section*{Funding information}
The authors extend their appreciation to the supported by Chongqing science and technology bureau (No. CSTB2024TIAD-CYKJCXX0009),
the Foundation of the Sichuan Provincial Department of Science and Technology (No. 2026NSFSC0138), the Key Laboratory of Numerical Simulation of Sichuan Provincial Universities (No. KLNS-2023SZFZ002 and KLNS-2023SZFZ005), and the Key Laboratory of Mathematical Meteorology (No. 2025Z0340).

\section*{Declaration of competing interest}
The authors declare that they have no known competing financial interests or personal relationships that could have appeared to influence the work reported in this paper.

\bibliographystyle{elsarticle-num} 
\bibliography{references}

\clearpage 
\appendix
\setcounter{table}{0}
\section{Complete experimental data results}
\label{Complete experimental results}
This appendix presents the complete, unsimplified mathematical expressions discovered by SGN across various physical systems and noise regimes. For the evaluation metrics in the subsequent tables, MSE(Noise) denotes the mean squared error evaluated on the corrupted data, whereas MSE(Original) measures the structural generalization error on the pristine ground truth. To facilitate interpretation of the raw expressions, which inherently adopt the programmatic naming conventions of our GNN architecture, Table~\ref{tab:nomenclature} details the physical and architectural meanings of all symbolic variables.

\begin{table}[htbp]
\centering
\small
\caption{Nomenclature and physical interpretations of the symbolic variables discovered by SGN.}
\label{tab:nomenclature}
\renewcommand{\arraystretch}{1.2}
\begin{tabular}{l p{12cm}}
\toprule
\textbf{Variable} & \textbf{Physical and Architectural Description} \\
\midrule
\texttt{t} & The temporal evolution step $t$ of the dynamical system. \\
\texttt{x}, \texttt{y} & The spatial coordinates of the grid nodes. \\
\texttt{u\_prev} & For scalar fields (Wave, Convection-Diffusion), the field value $u$ of the central node at the previous time step. \\
\texttt{x\_prev}, \texttt{y\_prev} & For vector fields (Navier-Stokes), the two state components of the central node at the previous time step, corresponding to horizontal ($u$) and vertical ($v$) velocities. \\
\texttt{*\_prev1} & In message functions, the historical state variable of the \textbf{neighbor node} (source node, $j$). \\
\texttt{*\_prev2} & In message functions, the historical state variable of the \textbf{central node} (target node, $i$). \\
\texttt{ux}, \texttt{uy} & The first-order spatial derivatives (gradients) of the field variable at the central node with respect to $x$ and $y$. \\
\texttt{uxx}, \texttt{uyy} & The second-order spatial derivatives (diffusion terms) of the field variable at the central node with respect to $x$ and $y$. \\
\texttt{ux1}, \texttt{uy1}, \dots & The corresponding first/second-order spatial derivatives evaluated at the \textbf{neighbor node}. \\
\texttt{ux2}, \texttt{uy2}, \dots & The corresponding first/second-order spatial derivatives evaluated at the \textbf{central node}. \\
\texttt{e0}, \texttt{e1} & The components of the aggregated low-dimensional \textbf{message vector} received by the central node, serving as inputs to the node update equation. \\
\bottomrule
\end{tabular}
\end{table}

\begin{table}[htbp]
\centering
\scriptsize
\caption{Unedited symbolic expressions discovered for the Wave Equation across varying noise levels.}
\label{tab:wave_full}
\renewcommand{\arraystretch}{1.3} 
\begin{tabular}{c c c c p{8.5cm}}
\toprule
\textbf{Noise} & \textbf{MSE(Noise)} & \textbf{MSE(Original)} & \textbf{Type} & \textbf{Equation} \\
\midrule

\multirow{2}{*}{0.0\%} & \multirow{2}{*}{/} & \multirow{2}{*}{2.5431E-04} 
& Msg  & sin(sin(sin(sin(cos(((t - t) + -0.56922716) + sin(((-1.8829098 - (t + ux2)) * -0.119090885) - u\_prev1))) + -1.2227476))) + -0.0041820854 \\
& & & Node & (-0.0049376567 + (sin(y * 3.0965688) * sin(x * 3.084605))) * cos(t / -0.31685314) \\
\midrule

\multirow{2}{*}{0.1\%} & \multirow{2}{*}{5.4111E-04} & \multirow{2}{*}{5.4102E-04} 
& Msg  & u\_prev2*(2.405846 - u\_prev1)*(u\_prev1*(-0.14979462) + ux2*(-0.017480928) - (-1)*0.02454977/(t - 1*0.5184039)) \\
& & & Node & sin(sin(x*3.1401417)*cos(t*(-3.1437826)))*sin(y + y + y/0.8408427)/sin(1.2261999) \\
\midrule

\multirow{2}{*}{0.2\%} & \multirow{2}{*}{4.7517E-07} & \multirow{2}{*}{1.3837E-10} 
& Msg  & ((cos(cos(sin(u\_prev1 + sin(u\_prev1 + u\_prev2)) + ((uy1 + (ux1 - t)) * 0.20625228))) * cos(t)) - (u\_prev2 - -0.08223844)) * ((u\_prev1 - 1.3154284) * -0.19282961) \\
& & & Node & sin(x*3.1416032)*sin(y*3.1416051)*cos(t*3.1416218) \\
\midrule

\multirow{2}{*}{0.5\%} & \multirow{2}{*}{2.9693E-06} & \multirow{2}{*}{1.4776E-10} 
& Msg  & (-u\_prev2 + sin(t*(u\_prev2*ux2 - 1*(-3.231915)))*0.06917196)*0.45004588 \\
& & & Node & sin(x*3.1415567)*sin(y*3.1416364)*cos(t/(-0.31830963)) \\
\midrule

\multirow{2}{*}{1.0\%} & \multirow{2}{*}{1.4262E-04} & \multirow{2}{*}{1.2472E-04} 
& Msg  & cos((uy1*0.025896823 + 0.096348494)*cos(t - (-0.31524143)*ux1) - (sin(u\_prev2)*0.66344243 + 0.48486096)) - 0.9074855 \\
& & & Node & u\_prev - (-0.06166134)*(0.008264106 - u\_prev)*sin(t*3.123437)/cos(t*3.086291) \\
\midrule

\multirow{2}{*}{2.0\%} & \multirow{2}{*}{1.2652E-04} & \multirow{2}{*}{3.8287E-05} 
& Msg  & t*(-0.03245001) - u\_prev2/(-1.6076635) + (u\_prev1 - 1*0.46895114)*(u\_prev1*0.23417574 + (-ux1 + uy1)*(-0.026761778) - 0.22886173) \\
& & & Node & u\_prev + sin(x*3.1438239)*(-0.06252605)*sin(t*3.1181076)* sin(y*3.1389296) \\
\midrule

\multirow{2}{*}{5.0\%} & \multirow{2}{*}{6.7079E-04} & \multirow{2}{*}{6.6427E-05} 
& Msg  & sin(-0.4441846*u\_prev2 + (2.4497662*u\_prev2*u\_prev2 + sin(t/0.32815966 - 1*0.65609646))*0.022165552) \\
& & & Node & ((((sin(t / 0.31949028) * -0.09182406) * sin(y / 0.3178091)) * 0.68839014) * sin(x / 0.3180929)) + (u\_prev / 0.99954045) \\
\midrule

\multirow{2}{*}{10.0\%} & \multirow{2}{*}{7.8418E-02} & \multirow{2}{*}{7.6029E-02} 
& Msg  & (u\_prev1 + (-sin(u\_prev1) - 2.8246412)*sin(u\_prev2))/(-6.849007) \\
& & & Node & u\_prev - sin((sin(t * -3.1070971) * sin(y * -3.1335773)) * 0.038306106) \\

\bottomrule
\end{tabular}
\end{table}

\begin{table}[htbp]
\centering
\scriptsize
\caption{Unedited symbolic expressions discovered for the Convection-Diffusion Equation across varying noise levels.}
\label{tab:cd_full}
\renewcommand{\arraystretch}{1.3} 
\begin{tabular}{c c c c p{8.5cm}}
\toprule
\textbf{Noise} & \textbf{MSE(Noise)} & \textbf{MSE(Original)} & \textbf{Type} & \textbf{Equation} \\
\midrule

\multirow{3}{*}{0.0\%} & \multirow{3}{*}{/} & \multirow{3}{*}{5.5649E-03} 
& Msg1 & (0.26722786 - 0.37149414*u\_prev2)*sin(u\_prev2*(ux1 + 2.8515506)) \\
& & & Msg2 & ux2*(sin(ux2)*0.04038851 + 0.02046432) \\
& & & Node & u\_prev + ((ux - ((u\_prev + (((uy * -0.08273945) - 0.50470436) * uy)) - (sin(u\_prev) * ((u\_prev * t) + -1.7263777)))) * -0.0040644426) \\
\midrule

\multirow{3}{*}{0.1\%} & \multirow{3}{*}{1.6789E-03} & \multirow{3}{*}{1.6792E-03} 
& Msg1 & (cos(uy2) - cos(uyy1*(-0.0684862)))*(cos(uyy2*(-0.075827084)) + 0.92147017)*(-0.10752371) \\
& & & Msg2 & u\_prev1*sin((u\_prev1 - 1*0.82263863)*sin(ux1 + 0.73160326) + sin(t*u\_prev2) + cos(u\_prev1)/0.27612534) \\
& & & Node & u\_prev + ((-u\_prev + ux*0.25640112)**2 - (ux - (-0.6112448)*uy))*0.0033418469 \\
\midrule

\multirow{3}{*}{0.5\%} & \multirow{3}{*}{1.4336E-03} & \multirow{3}{*}{1.4342E-03} 
& Msg1 & (-u\_prev1/1.3110251 + u\_prev2)*(t - cos(u\_prev1/0.1322431))*(-0.30173266) \\
& & & Msg2 & u\_prev1*(u\_prev2 - (-0.019017395)*(uyy1 - (-0.37160122)*(uxx1 - uyy2))) \\
& & & Node & u\_prev + (u\_prev - ux + uy*(-0.50821894))*0.0040541845 \\
\midrule

\multirow{3}{*}{1.0\%} & \multirow{3}{*}{3.0767E-03} & \multirow{3}{*}{3.0589E-03} 
& Msg1 & ((u\_prev1 + (u\_prev1 - (uyy1 * 0.007043898))) + ((ux2 + (uy1 * -0.43381253)) * 0.08615916)) * ((sin(u\_prev2 - (u\_prev1 * 1.4456807)) + ((t * 0.087506555) + 0.26573014)) * 0.9458101) \\
& & & Msg2 & sin(sin(sin(((sin(sin(u\_prev1) - t) * -0.38601163) * u\_prev1) - ((((-0.79170316 - u\_prev1) * u\_prev1) - (u\_prev2 * -1.593338)) * (sin(u\_prev2 / 0.68245786) * -1.6866558)))) \\
& & & Node & u\_prev - ((((e1 - square((ux + cos(uy)) - u\_prev)) * 0.049613487) + ((ux - u\_prev) + (uy / (sin((y + uy) * -0.3068668) + 2.3308575)))) * 0.0039940816) \\
\midrule

\multirow{3}{*}{2.0\%} & \multirow{3}{*}{2.3087E-03} & \multirow{3}{*}{2.3142E-03} 
& Msg1 & 0.010929662*(-ux1 + ux2) \\
& & & Msg2 & u\_prev1*(u\_prev1 - (t*0.08562517 + sin(u\_prev2)))/0.25098807 \\
& & & Node & ((((e1 + sin(uyy)) * 0.37931076) + (u\_prev - ((ux + uy) + ux))) * (0.0019810875 - ((ux - (uy + (uxx * -0.06388371))) * 0.000119407356))) + u\_prev \\
\midrule

\multirow{3}{*}{5.0\%} & \multirow{3}{*}{1.5740E-03} & \multirow{3}{*}{1.5162E-03} 
& Msg1 & (u\_prev1 * ((-0.4607998 - u\_prev1) + (t * 0.13325739))) + (u\_prev2 / 0.87254393) \\
& & & Msg2 & (((uxx2 * u\_prev1) + t) + sin(uy2)) * ((t * 0.0018317241) - ((sin(u\_prev2) - u\_prev1) * -0.022658467)) \\
& & & Node & u\_prev + ((((uy * y) - e1) + (((cos(ux) - exp(u\_prev * -107.679985)) + ux) - e1)) * -0.0040704776) \\
\midrule

\multirow{3}{*}{10.0\%} & \multirow{3}{*}{1.3019E-02} & \multirow{3}{*}{1.4153E-02} 
& Msg1 & u\_prev2 * (cos(sin(-0.037906747) * ((ux1 / -0.33668214) + uyy1)) * 0.39657518) \\
& & & Msg2 & uy2*(-0.007133365) - (-0.07567513)*(u\_prev2 - 0.0005025013*uyy1*uyy1) \\
& & & Node & u\_prev + ((ux + ((sin(u\_prev / 0.021259548) / (square(ux) + 0.21166894)) + -0.30247244)) * -0.0043281894) \\

\bottomrule
\end{tabular}
\end{table}

\begin{table}[htbp]
\centering
\scriptsize
\caption{Unedited symbolic expressions discovered for the Incompressible Navier-Stokes Equations across varying noise levels.}
\label{tab:ns_full}
\renewcommand{\arraystretch}{1.3} 
\begin{tabular}{c c c c p{8.5cm}}
\toprule
\textbf{Noise} & \textbf{MSE(Noise)} & \textbf{MSE(Original)} & \textbf{Type} & \textbf{Equation} \\
\midrule

\multirow{4}{*}{0.0\%} & \multirow{4}{*}{/} & \multirow{4}{*}{9.6899E-04} 
& Msg1  & -sin(t/(-6.233686) + y\_prev2) - 0.6374596 \\
& & & Msg2  & (cos((-0.118906446 * t) - cos(x\_prev1 + 0.21283843)) - y\_prev2) * 0.6010129 \\
& & & Node1 & sin(y - 1.5701683)*(ux + uy - sin(x))*cos(sin(t*(-0.1646834)) - 1*0.1230068)*0.9365822 \\
& & & Node2 & ((sin(t * -0.13451573) + 1.6071377) * sin(y)) * ((cos(x) - 0.007157283) * -0.6139257) \\
\midrule

\multirow{4}{*}{0.1\%} & \multirow{4}{*}{6.2724E-04} & \multirow{4}{*}{6.2718E-04} 
& Msg1  & (-x\_prev2 + y\_prev1)*0.015265973 \\
& & & Msg2  & 0.15940122 + (t - x\_prev2)*(-0.026457742) \\
& & & Node1 & ((e0 * -0.0015770864) + 0.9900558) * x\_prev \\
& & & Node2 & (0.9902582 - (e1 * 0.00073871564)) * y\_prev \\
\midrule

\multirow{4}{*}{0.5\%} & \multirow{4}{*}{6.3068E-04} & \multirow{4}{*}{6.2790E-04} 
& Msg1  & 0.23712291 + cos(-uy1 + x\_prev1 - y\_prev1)*(-0.21217735) \\
& & & Msg2  & (uy1 + y\_prev1)*(-0.004990364) \\
& & & Node1 & (e1 * -0.0046218093) - (x\_prev * -0.9900131) \\
& & & Node2 & (e1 * 0.0029781198) + (y\_prev * 0.9900553) \\
\midrule

\multirow{4}{*}{1.0\%} & \multirow{4}{*}{6.5369E-04} & \multirow{4}{*}{6.4082E-04} 
& Msg1  & (-x\_prev1 + x\_prev2)*(-0.07927439) \\
& & & Msg2  & (y\_prev2 / (t / -0.006975147)) + -0.0055887355 \\
& & & Node1 & (x\_prev * 0.9897041) + (e0 * -0.011039681) \\
& & & Node2 & ((e1 * -0.005140257) + y\_prev) * 0.9898316 \\
\midrule

\multirow{4}{*}{2.0\%} & \multirow{4}{*}{1.0362E-03} & \multirow{4}{*}{2.1334E-03} 
& Msg1  & (x\_prev1 - x\_prev2 + y\_prev1 - y\_prev2)*(-0.3086815) - 1*0.012407416 \\
& & & Msg2  & (-0.4978986*x\_prev1 - (-0.5724906)*x\_prev2 + y\_prev1 - 1.0975764*y\_prev2)/2.5570834 \\
& & & Node1 & (x\_prev * 0.9885945) - (e0 * -0.0075576548) \\
& & & Node2 & ((e1 * -0.0038489283) + y\_prev) * 0.98886806 \\
\midrule

\multirow{4}{*}{5.0\%} & \multirow{4}{*}{6.1238E-03} & \multirow{4}{*}{6.8773E-03} 
& Msg1  & (-y\_prev2 + torch.sin(x\_prev1 - x\_prev2 + y\_prev1))*0.5179765 - 1*0.01374288 \\
& & & Msg2  & (0.34813726 - uy1)*(-0.040036205) + (-x\_prev2 + y\_prev2 + (x\_prev1 - y\_prev1)*0.9260883)*(-0.5167467) \\
& & & Node1 & e1 + (((x\_prev - e1) * 0.98433083) - (e0 * 0.019087898)) \\
& & & Node2 & y\_prev - ((e0 + (y\_prev + torch.square(e1 + 0.27977818))) * 0.018425854) \\
\midrule

\multirow{4}{*}{10.0\%} & \multirow{4}{*}{3.9270E-02} & \multirow{4}{*}{3.9270E-02} 
& Msg1  & x\_prev2*0.34059486 + (x\_prev1 + y\_prev1*0.92205274 - y\_prev2)*(-0.27819878) - 1*0.008375836 \\
& & & Msg2  & (-x\_prev1 + y\_prev1 + (x\_prev2 - y\_prev2 + 0.02801064)*1.1195679)*0.37756348 \\
& & & Node1 & (e0*(-0.089349516) - x\_prev)*(-0.9576391) \\
& & & Node2 & y\_prev + (x*0.37457064 - (-t - 1.8729712)*(-e0 + e1*2.0220585 + y\_prev) - 1*1.8696856)*(-0.010119061) \\

\bottomrule
\end{tabular}
\end{table}

\end{document}